\documentclass{article}
\usepackage{spconf,graphicx,amsmath,pifont,amssymb}

\usepackage{subcaption}

\title{NaRLE: Natural Language Models using Reinforcement Learning with Emotion Feedback}

\name{Ruijie Zhou$^{\star}\thanks{This work is done during Ruijie's intern at Microsoft}$ \qquad Soham Deshmukh$^{\dagger}$ \qquad Jeremiah Greer$^{\dagger}$ \qquad Charles Lee$^{\dagger}$}
\address{$^{\star}$UC Berkeley $^{\dagger}$Microsoft \\ \texttt{ruijie@berkeley.edu, \{sdeshmukh,jegreer,charlle\}@microsoft.com}}
\begin{document}
%
\maketitle
\begin{abstract}
Current research in dialogue systems is focused on conversational assistants working on short conversations in either task-oriented or open domain settings. In this paper, we focus on improving task-based conversational assistants online, primarily those working on document-type conversations (e.g., emails) whose contents may or may not be completely related to the assistant's task. We propose ``NARLE" a deep reinforcement learning (RL) framework for improving the natural language understanding (NLU) component of dialogue systems online without the need to collect human labels for customer data. The proposed solution associates user emotion with the assistant's action and uses that to improve NLU models using policy gradients. For two intent classification problems, we empirically show that using reinforcement learning to fine tune the pre-trained supervised learning models improves performance up to 43\%. Furthermore, we demonstrate the robustness of the method to partial and noisy implicit feedback. 

 
\end{abstract}
\begin{keywords}
natural language understanding, deep reinforcement learning, emotion recognition, intent detection
\end{keywords}
\section{Introduction}
\label{sec:intro}
\begin{figure*}[ht] 
  \begin{subfigure}[b]{0.5\linewidth}
  \centering
  \includegraphics[width=.55\linewidth]{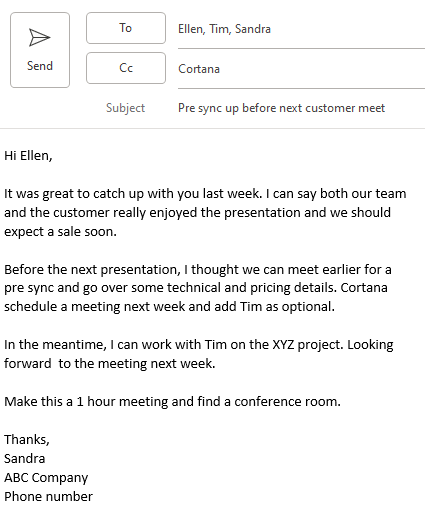}  
  \caption{Example email to document-type assistant}
  \label{fig:email_response.}
  \end{subfigure}
  \begin{subfigure}[b]{0.5\linewidth}
  \centering
  \centering
    \includegraphics[scale=0.15]{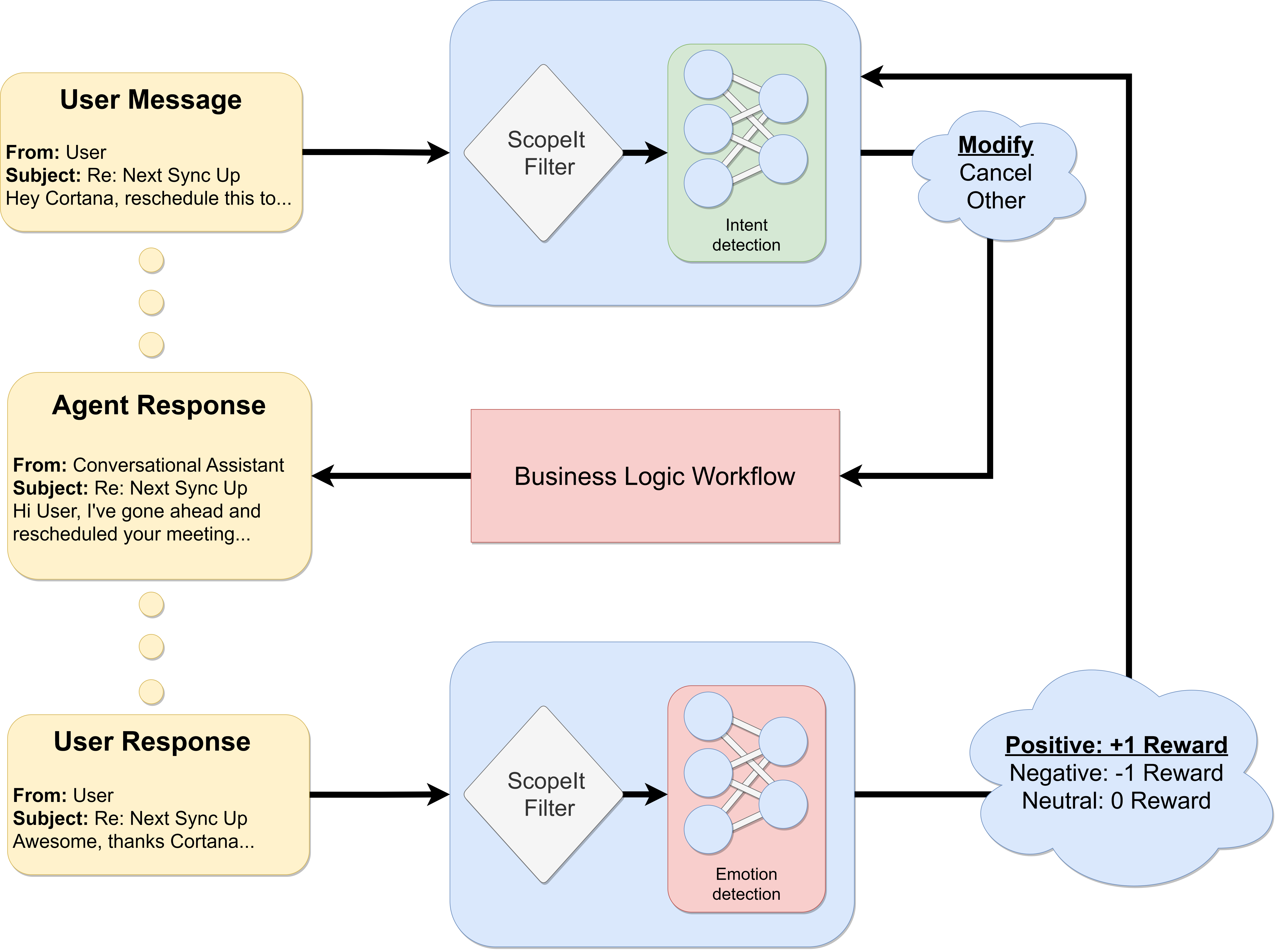}
    \caption{Architecture of proposed framework NARLE}
    \label{fig:RL_icassp}
  \end{subfigure}
  \caption{Improving natural language understanding using implicit feedback}
\end{figure*}

In recent years, Intelligent Personal Digital Assistants (IPDA) have becoming increasingly popular. Users interact with these assistants using natural language (either spoken or through text) where the conversations are generally short and chit-chat based. There is also emergence in assistants which interact with document-type conversation, such as emails (figure \ref{fig:email_response.}). These assistants must work with a broader context scope and multiple sub-conversations or intents occurring at each turn of the user input.
The increased scope and complexity that comes with multi-turn conversations creates many challenges for assistants in this setting. Specifically, extracting entities among other non-task entities, and ensuring dialogue state updates from entities relevant to the task become challenging. Previous works \cite{scopeit, htransformer} have shown that directly applying chit-chat methods to such document-type conversations tasks leads to sub-optimal results. 

In all conversation models, collecting and manually annotated training data is a challenge and incurs significant cost. This problem is exacerbated by the distributional shift of the input data in online decision-making, meaning a supervised learning model must be retrained periodically to maintain its accuracy. Moreover, for commercial assistants, most of the user data is not available for eyes-on access, either for collecting or labeling, and hence cannot be used for training or improving models via traditional supervised learning. In this paper, we ask and answer the question \textit{"Can we effectively improve assistants in an online setting without explicit feedback and no eyes-on access to the data?"}

In this paper, we provide a simple yet effective method to improve the NLU components of intelligent assistants in a privacy-preserving online setting. The framework consists of two main parts. In the first part, we associate a user's emotion in his/her response to the assistant with the assistant's previous action. This is critical in document-type conversations which have multiple intents present in each turn of a conversation where the emotion might not be associated to the task being completed by the assistant. In the second part, the relevant detected emotion is used as a weak reward signal for the policy gradient algorithm to update the NLU models online. The signal is associated with the previous intent/action the assistant took and is used to improve its behavior for the previous step. It's important to ensure users can provide feedback with minimal effort. We accomplish this by detecting the implicit emotion from the user's natural conversation with the assistant rather than requiring the user's explicit feedback (such as ratings which are difficult to collect and might not be reliable indicators on the assistant's performance).

\section{Related work}
\label{sec:related work}


Natural language understanding in conversations (including intent detection, entity extraction, and dialogue state tracking) are well studied problems \cite{gao2019neural, dst_review, todbert, simpletod}. Recently, there has been work in joint end-to-end models \cite{Liu+2016, Liu2017}. Jointly modeling multiple tasks in an end-to-end manner solves the model attribution problem and paves the way for directly optimizing the end-goal objective of correctly completing a task. This led to works which used a combination of supervised learning with RL \cite{li2017end-to-end, peng-etal-2017-composite, Dhingra2017EndtoEndRL}. Approaches which preferred pre-training the dialogue model before interactive tuning using RL suffered problems of dialogue state distribution mismatch. A solution to this was proposed by \cite{liu-etal-2018-dialogue}, who proposed a hybrid imitation and RL approach. The learning agent learned its policy from supervised training but used users' explicit guidance or demonstrations to learn the correct action. The feedback used in this case is either explicit feedback provided by user, or treating task completion as positive feedback.

One key component in the proposed framework is extracting emotion from document-type conversations by attributing the expressed emotion to the correct cause, as emotions not emerging from the assistant's actions should be ignored. Previous works focusing on emotion cause extraction and its variations evolve from rule-based methods \cite{chen2010emotion} to modern learning-based models \cite{gui2018event,xia2019emotion, chen2020conditional,poria2020recognizing}. Instead of relying on the above methods, we adapt the ScopeIt model \cite{scopeit} to extract not only task specific sentences but also sentences expressing emotion towards the tasks of interest to the assistant. Specifically, the ScopeIt is a neural model consisting of three parts: an intra-sentence aggregrator which contextualizes information within sentence, an inter-sentence aggregator which contextualizes information across sentences, and a classifier.

\section{Methodology}
This section contains the details of the proposed deep RL framework, with sections dedicated to the learning agent, the environment, and the learning algorithm. The overall architecture is depicted in Figure \ref{fig:RL_icassp}.

\subsection{The Learning Agent} \label{section:method:agent}
The learning agent in our case is a sequence of a ScopeIt unit \cite{scopeit} and a NLU model. The ScopeIt unit reduces the initial emails from the users to sentences relevant to the assistant's task, which will be supplied to the NLU model. In our work, the ScopeIt module is modified to identify both task specific sentences and sentences which provide surrounding information about the task. The agent will learn a policy that maps the filtered email message to the action based on the reward received from the environment. In this work, the learning agent can either start from scratch or from a pre-trained model using supervised learning.

\subsection{The Environment} \label{section:method:env}
The environment models the dynamics of users' interactions with the learning agent. Upon receiving the actions from the agent, the workflow will automatically generate a response based on the predicted action and the existing data and knowledge base. Next, when users respond to the agent's action, their response may express implicit emotion towards the action of the agents. Note that one key challenge in document-type conversations like emails is associating the expressed emotion to the correct cause, as emotions not emerging from the agent's actions should be ignored. We modify ScopeIt module to extract sentences expressing emotion towards the tasks of interest to the assistant. Filtered sentences are embedded using BERT, generating an emotion embedding. This emotion embedding is then used to classify the implicit emotion into positive, negative, or neutral. The detected emotion is then mapped to a numerical reward and fed back to the agent for updating the policy network.

\subsection{The Learning Algorithm} \label{section:method:alg} 
We optimize the NLU model by allowing the agent to interact with users and learn from user feedback. We only use implicit emotions associated with the task as the metric in designing the reward. A reward function $R(x)$ is collected respectively for each positive (x = 1), negative (x = -1) and neutral (x = 0) implicit feedback.

\[ R(x) = \begin{cases} 
         +1 & x = 1 \\
         -1 & x = -1 \\
         0 & x = 0 
       \end{cases}
    \]

The agent always acts on-policy in order to ensure the agent is always acting optimally for customers. This is accomplished through a softmax policy on the actions/predictions from the policy network outputs.  We apply the REINFORCE algorithm \cite{williams1992simple} in optimizing the network parameters. Since the expectation of the sample gradient is equal to the actual gradient, we measure returns from real sample trajectories and use that to update our policy gradient. Specifically,
\begin{equation*}
    \begin{aligned}
   \nabla_{\theta}J(\theta) & = \mathbb{E}_{\pi}[Q^{\pi}(s,a) \nabla_{\theta}\ln(\pi_{\theta}(a|s))] \\
   & =  \mathbb{E}_{\pi}[\nabla_{\theta}\ln(\pi_{\theta}(a|s)) R]
   \end{aligned}
\end{equation*}
where $Q^{\pi}(s,a)$ is the value of state-action pair when we follow a policy $\pi$.

\section{Experiments}
\subsection{Dataset}\label{sec:exp}

The emotion detection model is trained offline on a private customer dataset from our private preview service. The dataset for emotion recognition model is built from the email messages that are either filtered or not filtered by the ScopeIt module, and user feedback directed toward the actions of the agents. The dataset is constructed through the following two steps. 

\textit{Step 1: Identifying insertion positions} We parse all email sentences based on a set of punctuation (including comma, period, colon, question and exclamation mark). Then, all positions after those punctuation are considered as candidate locations for injecting emotional feedbacks. 

\textit{Step 2: Inject users' emotions} We next inject the users' emotions at the candidate locations found in Step 1 randomly and label those samples. We also randomly insert ”general positive or negative emotions” and label those samples as neutral. This allows the agent to capture the nuance between emotions directed toward the agent and general emotions. 

\subsection{Setup}\label{sec:setup}
We use the following transformer-based models for the emotion detection problem: BERT, DistilBERT, ALBERT, and RoBERTa. For each model, a linear layer is added on top of the pooled output to perform classification. We freeze some transformer layers of all BERT-type models and the number of frozen layers are learned in a trial-and-error manner. The model was end-to-end trained using huggingface implementation \cite{wolf2019huggingface} on 4 Nvidia K80 GPUs.


\section{Results}\label{sec:res}
\subsection{Emotion recognition} 

We report both accuracy and the overall Macro F1 scores for all models in Table \ref{tab:res}. It is worth noting that the use of the ScopeIt model leads to improved performance for all models. Both the BERT and the DistilBERT models have similar accuracy. Hence, DistilBERT with two frozen layers is finally selected due to its smaller size and lower inference latency.  
\begin{table}[!t]
\small
\begin{tabular}{ccccc}
\hline
Models & ScopeIt & F1 & Accuracy & Parameters \\ \hline
ALBERT &  & 89.93 & 90.36 & 11M \\
RoBERTa(-2) &  & 91.22 & 91.58 & 125M \\
BERT(-1)       &  & 94.06          & 94.37          & 110M              \\
DistilBERT(-1) &  & 94.15          & 94.45          & 66M               \\
ALBERT     & \checkmark & 91.46          & 91.87          & 11M               \\
RoBERTa(-2)    & \checkmark & 91.75          & 92.03          & 125M              \\
BERT(-3)       & \checkmark & \textbf{94.96}          & \textbf{95.42}          & 110M              \\
DistilBERT(-2) & \checkmark & 94.92 & 95.39 & \textbf{66M}               \\ \hline
\end{tabular}
\caption{Results for emotion recognition model. All scores are in percentage and are reported at best accuracy. BERT(-1) represents the BERT classification model with one frozen transformer layer.}
\label{tab:res}
\end{table}

\begin{figure*}[ht] 
  \begin{subfigure}[b]{0.5\linewidth}
  \centering
  \includegraphics[width=.9\linewidth]{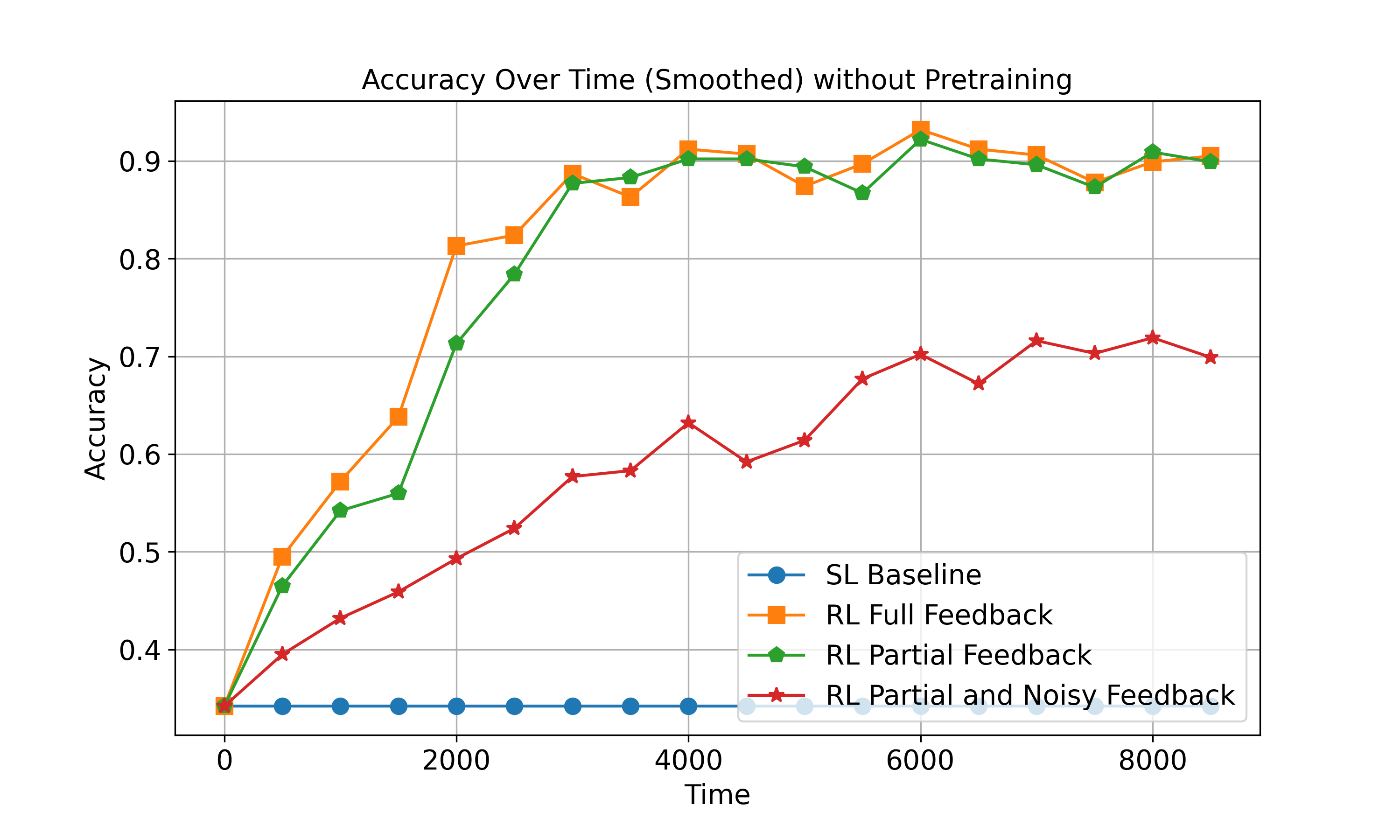}  
  \caption{Multi-class training on random model}
  \label{fig:MC_no_pretrain}
  \end{subfigure}
  \begin{subfigure}[b]{0.5\linewidth}
  \centering
  \includegraphics[width=0.9\linewidth]{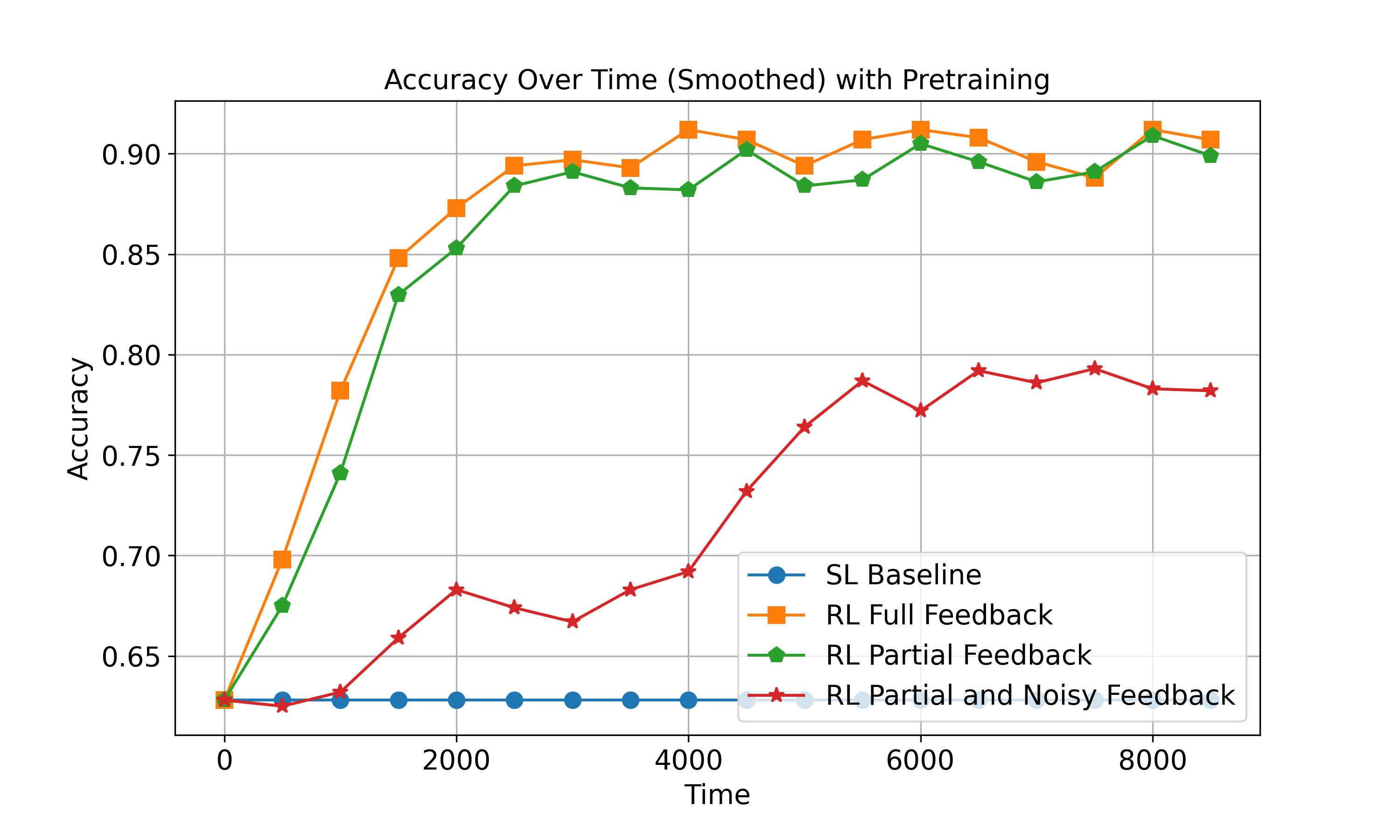}  
  \caption{Multi-class training on fine-tuned model}
  \label{fig:MC_pretrain}
  \end{subfigure} 
  \begin{subfigure}[b]{0.5\linewidth}
  \centering
  \includegraphics[width=.9\linewidth]{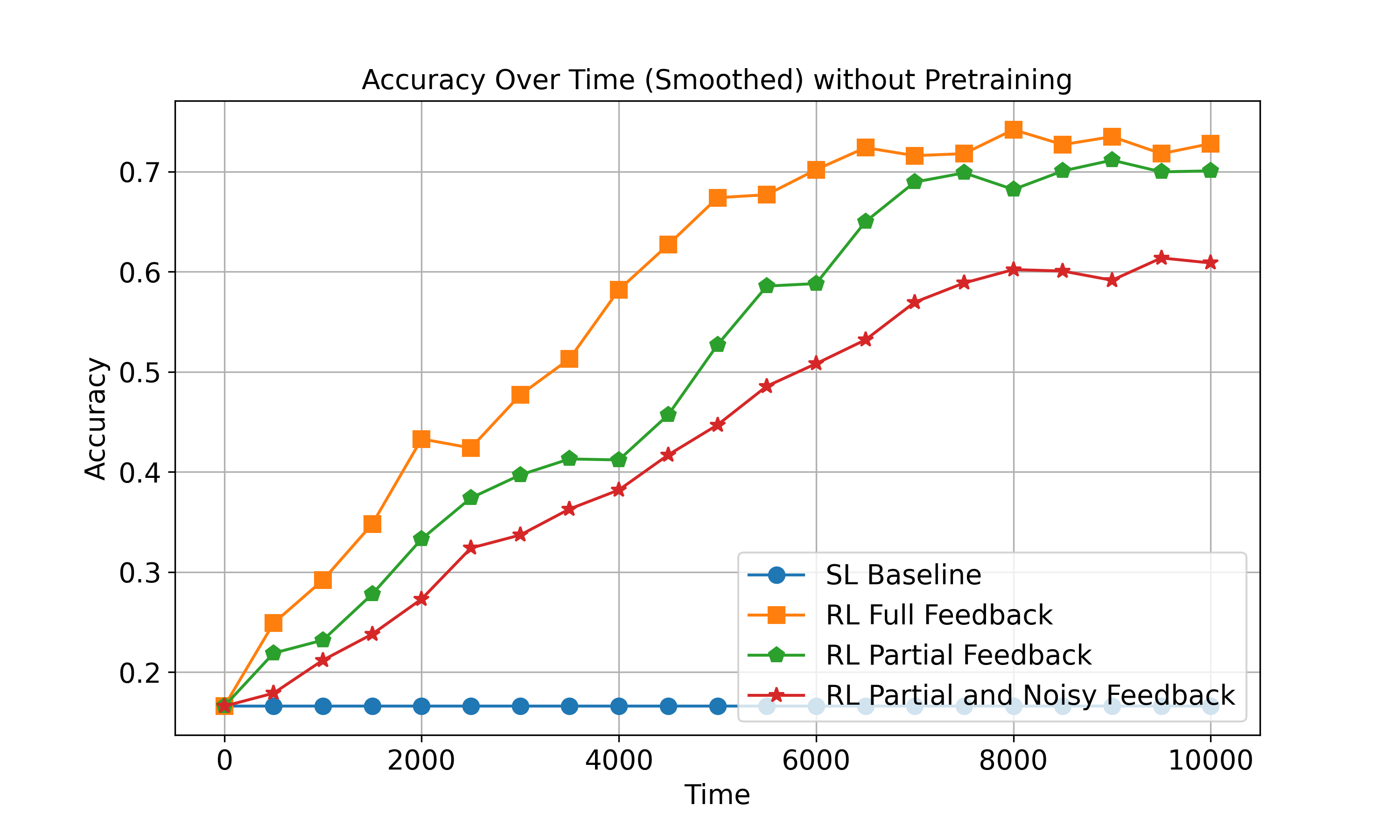}  
  \caption{Multi-label training on random model}
  \label{fig:ML_no_pretrain}
  \end{subfigure}
  \begin{subfigure}[b]{0.5\linewidth}
  \centering
  \includegraphics[width=.9\linewidth]{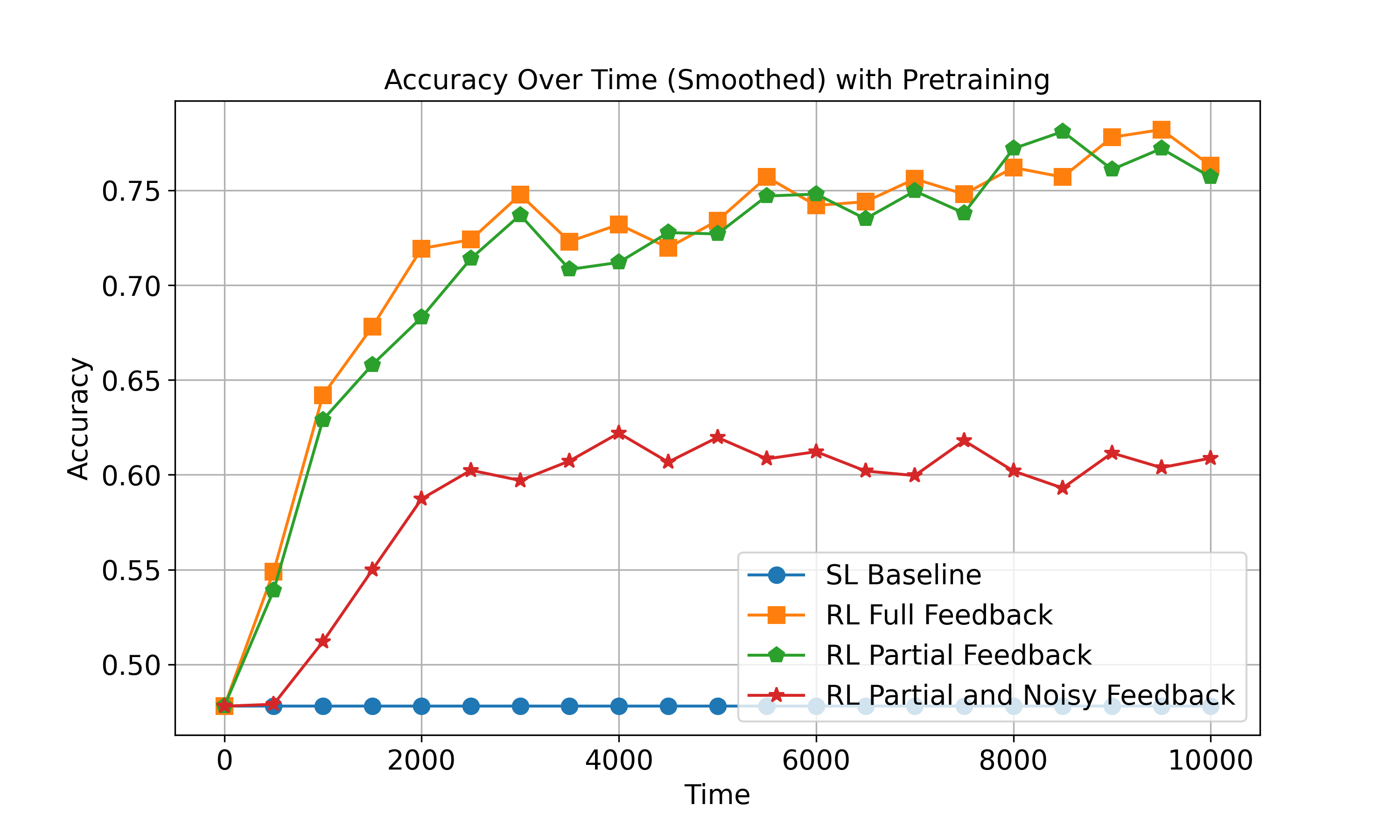}  
  \caption{Multi-label training on fine-tuned model}
  \label{fig:ML_pretrain}
  \end{subfigure} 
  \caption{Learning curves for different model and settings}
  \label{fig7} 
\end{figure*}

\subsection{Multi-class Intent Detection} 
We first consider an intent detection model behind a conversational assistant that assists users to schedule meetings. The assistant will use a DistilBERT model to classify the intent of emails into three categories: modify a meeting, cancel a meeting, and other (not relevant to task of meeting scheduling). We first conduct experiments in learning the intent classification model from scratch using only RL and summarize the results in figure \ref{fig:MC_no_pretrain}. We can see that all DistilBERT models start from random guesses, but as the learning agent interacts more with users the task success rates improve over time. 

The type of feedback obtained is critical to NLU model training. Hence, three different feedback mechanisms are studied in this work: full feedback, partial feedback, and partial with noisy feedback. In the full feedback scenario, every customer is assumed to leave implicit feedback. In partial feedback scenario only 15\% of requests are assumed to have implicit feedback. While in partial with noisy feedback, out of the 15\% partial feedback, one third of the implicit feedback is incorrect. These feedback scenarios provide insight into the NLU model performance and both quantity and quality of feedback required.

The orange curve shows the performance by assuming every customer leaves an implicit feedback. Even under partial (i.e., 15\%) feedback, the agent can achieve comparable performance with the full feedback case after sufficient number of turns. Finally, the agent learns more slowly and has lower accuracy under the partial and noisy feedback case where one third of the 15\% feedback are wrongly labeled.

We next demonstrate the experimental results that use online RL training for a limited data fine-tuned DistilBERT model. As shown in the blue line in figure \ref{fig:MC_pretrain}, the supervised benchmark model only has about 63\% accuracy. This is due to limited eyes-on access to data and the domain mismatch of offline training and online product scale-up. We study the same three feedback scenarios as before. Note that a supervised pre-trained model leads to faster learning rates for both full and partial feedback cases and a better accuracy for the partial and noisy feedback.



  




\subsection{Multi-label Intent Detection} 
We consider a multi-label intent classification model of the same conversational assistant as before. In this case, there are a total of six distinct actions and the action space is represented by a six-dimension vector with 0s and 1s. Out of the total 64 ($2^{6}=64$) possible scenarios, only six possible combinations are valid. The learning agent consists of an ensemble of six separate DistilBERT binary classification models which work independently to determine whether each single action should be taken or not. 

We first train the learning agent from scratch where all six DistilBERT models start from random guesses. The learning curves on accuracy for full, partial as well as partial and noisy cases are shown in figure \ref{fig:ML_no_pretrain}. We can see that the learning accuracy improves for all three scenarios over time. Specifically, the accuracy for both full and partial feedback can reach about 70\%, which is nearly 10\% higher than that of the partial noisy case. 

We next use online RL training for the six separate fine-tuned DistilBERT models from limited data. The learning curves are demonstrated in figure \ref{fig:ML_pretrain}, where the fine-tuned benchmark can be considerably improved over iterations for all scenarios. Also, compared to learning from scratch, fine-tuned models have faster learning rates and higher accuracies.






\section{Conclusion}\label{sec:conclusion}
We propose a deep RL framework ``NARLE" to improve the NLU models of a task-oriented conversational assistant in an online manner for document-type conversations. The proposed architecture scopes out emotion feedback relevant to the assistant's task and uses that as feedback to improve the performance of the assistant in an adaptive way. The proposed framework is evaluated on customer data, where the proposed method can improve a limited data fine-tuned model up to 43\%. We also show that the proposed method is robust to partial and noisy feedback. 



\bibliographystyle{IEEEbib}
\bibliography{strings}

\begin{thebibliography}{10}

\bibitem{scopeit}
Barun Patra, Vishwas Suryanarayanan, Chala Fufa, Pamela Bhattacharya, and
  Charles Lee,
\newblock ``\{S\}cope\{I\}t: Scoping task relevant sentences in documents,''
\newblock in {\em Proceedings of the 28th International Conference on
  Computational Linguistics: Industry Track}, Online, Dec. 2020, pp. 214--227,
  International Committee on Computational Linguistics.

\bibitem{htransformer}
Raghavendra Pappagari, Piotr Zelasko, Jesús Villalba, Yishay Carmiel, and
  Najim Dehak,
\newblock ``Hierarchical transformers for long document classification,''
\newblock in {\em 2019 IEEE Automatic Speech Recognition and Understanding
  Workshop (ASRU)}, 2019, pp. 838--844.

\bibitem{gao2019neural}
Jianfeng Gao, Michel Galley, and Lihong Li,
\newblock ``Neural approaches to conversational ai,''
\newblock {\em Foundations and Trends in Information Retrieval}, February 2019.

\bibitem{dst_review}
Matthew Henderson,
\newblock ``Machine learning for dialog state tracking: A review,''
\newblock in {\em Proceedings of The First International Workshop on Machine
  Learning in Spoken Language Processing}, 2015.

\bibitem{todbert}
Chien-Sheng Wu, Steven~C.H. Hoi, Richard Socher, and Caiming Xiong,
\newblock ``{TOD}-{BERT}: Pre-trained natural language understanding for
  task-oriented dialogue,''
\newblock in {\em Proceedings of the 2020 Conference on Empirical Methods in
  Natural Language Processing (EMNLP)}, Online, Nov. 2020, pp. 917--929,
  Association for Computational Linguistics.

\bibitem{simpletod}
Ehsan Hosseini-Asl, Bryan McCann, Chien-Sheng Wu, Semih Yavuz, and Richard
  Socher,
\newblock ``A simple language model for task-oriented dialogue,''
\newblock in {\em Advances in Neural Information Processing Systems},
  H.~Larochelle, M.~Ranzato, R.~Hadsell, M.~F. Balcan, and H.~Lin, Eds. 2020,
  vol.~33, pp. 20179--20191, Curran Associates, Inc.

\bibitem{Liu+2016}
Bing Liu and Ian Lane,
\newblock ``Attention-based recurrent neural network models for joint intent
  detection and slot filling,''
\newblock in {\em Interspeech 2016}, 2016, pp. 685--689.

\bibitem{Liu2017}
Bing Liu and Ian Lane,
\newblock ``An end-to-end trainable neural network model with belief tracking
  for task-oriented dialog,''
\newblock in {\em Proc. Interspeech 2017}, 2017, pp. 2506--2510.

\bibitem{li2017end-to-end}
Xiujun Li, Yun-Nung Chen, Lihong Li, Jianfeng Gao, and Asli Celikyilmaz,
\newblock ``End-to-end task-completion neural dialogue systems,''
\newblock in {\em the 8th International Joint Conference on Natural Language
  Processing}. November 2017, IJCNLP 2017.

\bibitem{peng-etal-2017-composite}
Baolin Peng, Xiujun Li, Lihong Li, Jianfeng Gao, Asli Celikyilmaz, Sungjin Lee,
  and Kam-Fai Wong,
\newblock ``Composite task-completion dialogue policy learning via hierarchical
  deep reinforcement learning,''
\newblock in {\em Proceedings of the 2017 Conference on Empirical Methods in
  Natural Language Processing}, Copenhagen, Denmark, Sept. 2017, pp.
  2231--2240, Association for Computational Linguistics.

\bibitem{Dhingra2017EndtoEndRL}
Bhuwan Dhingra, Lihong Li, Xiujun Li, Jianfeng Gao, Yun-Nung Chen, Faisal
  Ahmed, and L.~Deng,
\newblock ``End-to-end reinforcement learning of dialogue agents for
  information access,''
\newblock in {\em ACL}, 2017.

\bibitem{liu-etal-2018-dialogue}
Bing Liu, Gokhan T{\"u}r, Dilek Hakkani-T{\"u}r, Pararth Shah, and Larry Heck,
\newblock ``Dialogue learning with human teaching and feedback in end-to-end
  trainable task-oriented dialogue systems,''
\newblock in {\em Proceedings of the 2018 Conference of the North {A}merican
  Chapter of the Association for Computational Linguistics: Human Language
  Technologies, Volume 1 (Long Papers)}, June 2018, pp. 2060--2069.

\bibitem{chen2010emotion}
Ying Chen, Sophia Yat~Mei Lee, Shoushan Li, and Chu-Ren Huang,
\newblock ``Emotion cause detection with linguistic constructions,''
\newblock in {\em Proceedings of the 23rd International Conference on
  Computational Linguistics (Coling 2010)}, 2010, pp. 179--187.

\bibitem{gui2018event}
Lin Gui, Ruifeng Xu, Dongyin Wu, Qin Lu, and Yu~Zhou,
\newblock ``Event-driven emotion cause extraction with corpus construction,''
\newblock in {\em Social Media Content Analysis: Natural Language Processing
  and Beyond}, pp. 145--160. World Scientific, 2018.

\bibitem{xia2019emotion}
Rui Xia and Zixiang Ding,
\newblock ``Emotion-cause pair extraction: A new task to emotion analysis in
  texts,''
\newblock {\em arXiv preprint arXiv:1906.01267}, 2019.

\bibitem{chen2020conditional}
Xinhong Chen, Qing Li, and Jianping Wang,
\newblock ``Conditional causal relationships between emotions and causes in
  texts,''
\newblock in {\em Proceedings of the 2020 Conference on Empirical Methods in
  Natural Language Processing (EMNLP)}, 2020, pp. 3111--3121.

\bibitem{poria2020recognizing}
Soujanya Poria, Navonil Majumder, Devamanyu Hazarika, Deepanway Ghosal, Rishabh
  Bhardwaj, Samson Yu~Bai Jian, Romila Ghosh, Niyati Chhaya, Alexander Gelbukh,
  and Rada Mihalcea,
\newblock ``Recognizing emotion cause in conversations,''
\newblock {\em arXiv preprint arXiv:2012.11820}, 2020.

\bibitem{williams1992simple}
Ronald~J Williams,
\newblock ``Simple statistical gradient-following algorithms for connectionist
  reinforcement learning,''
\newblock {\em Machine learning}, vol. 8, no. 3, pp. 229--256, 1992.

\bibitem{wolf2019huggingface}
Thomas Wolf, Lysandre Debut, Victor Sanh, Julien Chaumond, Clement Delangue,
  Anthony Moi, Pierric Cistac, Tim Rault, R{\'e}mi Louf, Morgan Funtowicz,
  et~al.,
\newblock ``Huggingface's transformers: State-of-the-art natural language
  processing,''
\newblock {\em arXiv preprint arXiv:1910.03771}, 2019.

\end{thebibliography}

\end{document}